\documentclass[10pt,twocolumn,letterpaper]{article}

\usepackage{cvpr}              % To produce the CAMERA-READY version
\usepackage{graphicx}
\usepackage{amsmath}
\usepackage{amssymb}
\usepackage{booktabs}

\usepackage{enumitem}
\setlist{leftmargin=4.5mm}

\newcommand{\vggf}{\textsc{VGGface 2}}
\newcommand{\ds}{\textsc{DeepSpectrum}}

\usepackage[breaklinks,colorlinks]{hyperref}

\usepackage[capitalize]{cleveref}
\crefname{section}{Sec.}{Secs.}
\Crefname{section}{Section}{Sections}
\Crefname{table}{Table}{Tables}
\crefname{table}{Tab.}{Tabs.}

\def\confName{CVPR}
\def\confYear{2022}

\begin{document}

%%%%%%%%% TITLE - PLEASE UPDATE
\title{ABAW: Valence-Arousal Estimation, Expression Recognition, Action Unit Detection \& Emotional Reaction Intensity Estimation Challenges}

\author{Dimitrios Kollias
\and
Panagiotis Tzirakis
%Hume AI, USA
\and
Alice Baird
%Hume AI, USA
\and
Alan Cowen
%Hume AI, USA
\and
Stefanos Zafeiriou}
%Imperial College London, UK
%{\tt\small s.zafeiriou@imperial.ac.uk}

\maketitle

%%%%%%%%% ABSTRACT
\begin{abstract}

The fifth Affective Behavior Analysis in-the-wild (ABAW) Competition is part of the respective ABAW Workshop which will be held in conjunction with IEEE Computer Vision and Pattern Recognition Conference (CVPR), 2023. The 5th ABAW Competition is a continuation of the Competitions held at ECCV 2022, IEEE CVPR 2022, ICCV 2021, IEEE FG 2020 and CVPR 2017 Conferences, and is dedicated at automatically analyzing affect. 
For this year’s Competition, we feature two corpora: i) an extended version of the Aff-Wild2 database and ii) the Hume-Reaction dataset. The former database is an audiovisual one of around 600 videos of around 3M frames and is annotated with respect to: a) two continuous affect dimensions -valence (how positive/negative a person is) and arousal (how active/passive a person is)-; b) basic expressions (e.g. happiness, sadness, neutral state); and c) atomic facial muscle actions (i.e., action units). The latter dataset is an audiovisual one in
which reactions of individuals to emotional stimuli have been annotated with respect to seven emotional expression intensities.
Thus the 5th ABAW Competition encompasses four Challenges: i) uni-task Valence-Arousal Estimation, ii) uni-task Expression Classification, iii) uni-task Action Unit Detection, and iv) Emotional Reaction Intensity Estimation.
In this paper, we present these Challenges, along with their corpora, we outline the evaluation metrics, we present the baseline systems and  illustrate their obtained performance.

\end{abstract}

%%%%%%%%% BODY TEXT
\section{Introduction}

The ABAW Workshop and Competition has a unique aspect of fostering cross-pollination of different disciplines, bringing together experts (from academia, industry, and government) and researchers of mobile and ubiquitous computing, computer vision and pattern recognition, artificial intelligence and machine learning, multimedia, robotics, HCI, ambient intelligence and psychology. The diversity of human behavior, the richness of multi-modal data that arises from its analysis, and the multitude of applications that demand rapid progress in this area ensure that our events provide a timely and relevant discussion and dissemination platform.

The ABAW Workshop tackles the problem of affective behavior analysis in-the-wild, that is a major targeted characteristic of HCI systems used in real life applications. The target is to create machines and robots that are capable of understanding people's feelings, emotions and behaviors; thus, being able to interact in a 'human-centered' and engaging manner with them, and effectively serving them as their digital assistants. This interaction should not be dependent on the respective context, nor the human's age, sex, ethnicity, educational level, profession, or social position. As a result, the development of intelligent systems able to analyze human behaviors in-the-wild can contribute to generation of trust, understanding and closeness between humans and machines in real life environments.

The ABAW Workshop includes the respective Competition which utilizes two corpora: i) an extended version of the Aff-Wild2 database \cite{zafeiriou2017aff,kollias2017recognition,kollias2018deep1,kollias2019expression,kollias2020analysing,kollias2021analysing,kollias2021affect,kollias2021distribution,kollias2022abaw,kollias2023abaw,kollias2019face} and ii) the Hume-Reaction dataset. Aff-Wild2 database is an audiovisual one consisting of around 600 videos of around 3M frames and is annotated with respect to three different models of affect: a) dimensional affect (valence, which characterises an emotional state on a scale from positive to negative, and arousal, which characterises an emotional state on a scale from active to passive); b) categorical affect (six basic expressions -anger, disgust, fear, happiness, sadness, surprise- plus the neutral state); and c) action units (i.e., activations of facial muscles).
The Hume-Reaction dataset is an audiovisual one in
which reactions of individuals to emotional stimuli have been annotated with respect to seven emotional expression intensities (i.e., adoration, amusement, anxiety, disgust, empathic pain, fear and surprise).

Using these introduced datasets, the 5th ABAW Competition addresses four contemporary affective computing problems: in the Valence-Arousal Estimation Challenge, valence and arousal have to be predicted; in the Expression Classification Challenge, 6 basic expressions, the neutral state and a category 'other' (that denotes affective states that do not belong to the categorical model of affect) have to be recognised; in the Action Unit Detection Challenge, 12 action units (AUs) have to be detected; in the Emotional Reaction Intensity Estimation Challenge, seven fine-grained ‘in-the-wild’ emotions have to be predicted. The 3 former Challenges are based on the Aff-Wild2 database, whereas the latter Challenge is based on the Hume-Reaction dataset.

By providing the mentioned tasks in the 5th ABAW Competition, we aim to address research questions that are of interest to affective
computing, machine learning and multimodal signal processing
communities and encourage a fusion of their disciplines.

The fifth ABAW Competition, held in conjunction with the IEEE Computer Vision and
Pattern Recognition Conference (CVPR), 2023 is a continuation of the series of ABAW Competitions held in conjunction with ECCV 2022, IEEE CVPR 2022, ICCV 2021, IEEE FG 2020 and IEEE CVPR 2017.

\section{Competition Corpora}\label{corpora}

In the following, we provide a short overview of each Challenge's dataset. %and refer the reader to the original work for a more complete description.
Finally, we describe the pre-processing steps that we carried out for cropping and aligning all provided images. The cropped and aligned images have been utilized in our baseline experiments. 

\subsection{Valence-Arousal Estimation Challenge}

This Challenge's corpora include $594$ videos (an augmented version of the Aff-Wild2 database) that contain annotations in terms of valence and arousal. Sixteen of these videos display two subjects, both of which have been annotated. In total,  around 3 million frames, with $584$ subjects % posa frames akrivws k posa male k female
have been annotated by four experts using the method proposed in \cite{cowie2000feeltrace}. Valence and arousal values range continuously in $[-1,1]$. %Figure \ref{va_annot} shows the 2D Valence-Arousal histogram of annotations.

%\begin{figure}[h]
%\centering
%\includegraphics[height=5.6cm]{hist_va.png}
%\caption{Valence-Arousal Estimation Challenge: 2D Valence-Arousal Histogram of Annotations in Aff-Wild2}
%\label{va_annot}
%\end{figure}

Aff-Wild2 is split into training, validation and testing sets. Partitioning is done in a subject independent manner, in the sense that a person can  appear  strictly  in  only one  of  these sets. %These sets consist of 341, 71 and 152 videos, respectively. 

\subsection{Expression Classification Challenge}

%%%%
This Challenge's corpora include $546$ videos in Aff-Wild2 that contain annotations in terms of the the 6 basic expressions, plus the neutral state, plus a category 'other' that denotes expressions/affective states other than the 6 basic ones. Seven of these videos display two subjects, both of which have been annotated. In total, $2,624,160$ frames, with $437$ subjects, $268$ of which are male and $169$ female, have been annotated by seven experts in a frame-by-frame basis. 
%
%Due to subjectivity of annotators and wide ranging levels of images’ difficulty, there were some disagreements among annotators. We decided to keep only the annotations on which at least six (out of seven) experts agreed. 
Table \ref{expr_distr} shows the distribution of the expression annotations of Aff-Wild2.

\begin{table}[!h]
\caption{Expression Classification Challenge: Number of Annotated Images for each Expression  }
\label{expr_distr}
\centering
\begin{tabular}{ |c||c| }
\hline
 Expressions & No of Images \\
\hline
\hline
Neutral & 468,069  \\
 \hline
Anger & 36,627  \\
 \hline
Disgust & 24,412 \\
 \hline
Fear &  19,830 \\
 \hline
Happiness & 245,031  \\
 \hline
Sadness & 130,128  \\
 \hline
Surprise & 68,077  \\
 \hline
 Other & 512,262 \\
 \hline
\end{tabular}
\end{table} 
 
Aff-Wild2 is split into training, validation and testing sets, in a subject independent manner. %These sets consist of 248, 70 and 228 videos, respectively.

\subsection{Action Unit Detection Challenge}

This Challenge’s corpora include 541 videos that contain annotations in terms of 12 AUs, namely AU1, AU2, AU4, AU6, AU7, AU10, AU12, AU15, AU23, AU24, AU25 and AU26. Seven of these videos display two subjects, both of which have been annotated. In total, $2,627,632$ frames,with $438$ subjects, $268$ of which are male and $170$ female, have been annotated in a semi-automatic procedure (that involves manual and automatic annotations). The annotation has been performed in a frame-by-frame basis. Table \ref{au_distr} shows the name of the twelve action units that have been annotated, the action that they are associated with and the distribution of their annotations in Aff-Wild2.

\begin{table}[h]
%%% \vskip-0.15cm
    \centering
        \caption{Action Unit Detection Challenge: Distribution of AU Annotations in Aff-Wild2}
    \label{au_distr}
\begin{tabular}{|c|c|c|}
\hline
  Action Unit \# & Action   &\begin{tabular}{@{}c@{}} Total Number \\  of Activated AUs \end{tabular} \\   \hline    
    \hline    
   AU 1 & inner brow raiser   & 301,102 \\   \hline    
   AU 2 & outer brow raiser  & 139,936 \\   \hline   
   AU 4 & brow lowerer   & 386,689  \\  \hline    
   AU 6 & cheek raiser  & 619,775 \\  \hline    
   AU 7 & lid tightener  & 964,312 \\  \hline    
   AU 10 & upper lip raiser  & 854,519 \\  \hline    
   AU 12 & lip corner puller  & 602,835 \\  \hline    
   AU 15 & lip corner depressor  & 63,230 \\  \hline   
  AU 23 & lip tightener & 78,649 \\  \hline    
   AU 24 & lip pressor & 61,500 \\  \hline    
   AU 25 & lips part  & 1,596,055 \\  \hline     
   AU 26 & jaw drop  & 206,535 \\  \hline     
\end{tabular}
\end{table}

Aff-Wild2 is split into training, validation and testing sets, in a subject independent manner. %These sets consist of 295, 105 and 141 videos, respectively.

 \subsection{Emotional Reaction Intensity Estimation Challenge}

For the Emotional Reaction Intensity Estimation Challenge Challenge, the large-scale and in-the-wild Hume Reaction dataset is used. The participants of this sub-challenge explore a multi-output regression task, utilizing seven, self-annotated, nuanced classes of emotion: `Adoration,' `Amusement,' `Anxiety,' `Disgust,' `Empathic-Pain,' `Fear,' and `Surprise.' The dataset is multimodal, and the video samples were recorded in uncontrolled environmental conditions in a wide variety of at-home recording settings with varying background and lightning noise conditions. In total, $2,222$ participants from two cultures, South Africa ($1,084$) and the United States ($1,138$), aged from $18.5$ – $49.0$ years old, recorded their facial and vocal reactions to a wide range of emotionally evocative videos via their webcam. 

\subsection{Aff-Wild2 Pre-Processing: Cropped \& Cropped-Aligned Images} \label{pre-process}

%In order to help the participating teams, we performed: i) cropping on all the images (videoframes) of Aff-Wild2 (after splitting all videos into frames) and ii) cropping on all the images and aligning them.

At first, all videos are splitted into independent frames. Then they are passed through the RetinaFace detector.
so as to extract, for each frame, face bounding boxes and 5 facial landmarks. The images were cropped according the bounding box locations; then the images were provided to the participating teams.
The 5 facial landmarks (two eyes, nose and two mouth corners) were  used to perform similarity transformation. The resulting cropped and aligned images were additionally provided to the participating teams. Finally, the cropped and aligned images were utilized in our baseline experiments, described in Section \ref{baseline}.

All   cropped   and   cropped-aligned   images   were  resized   to $112 \times 112 \times 3$ pixel resolution and their intensity values were normalized  to  $[-1,1]$.

\section{Evaluation Metrics Per Challenge}\label{metrics}

%Next, we present the metrics that will be used for assessing the performance of the developed methodologies of the participating teams in each Challenge.

\subsection{Valence-Arousal Estimation Challenge}

The performance measure is the average between the Concordance Correlation Coefficient (CCC) of valence and arousal:

\begin{equation} \label{va}
\mathcal{P}_{VA} = \frac{CCC_a + CCC_v}{2}
\end{equation}

CCC evaluates the agreement between two time series (e.g., all video annotations and predictions) by scaling their correlation coefficient with their mean square difference. CCC takes values in the range $[-1,1]$; high values are desired. CCC is defined as follows:

\begin{equation} \label{ccc}
CCC = \frac{2 s_{xy}}{s_x^2 + s_y^2 + (\bar{x} - \bar{y})^2},
\end{equation}

\noindent
where $s_x$ and $s_y$ are the variances of all video valence/arousal annotations and predicted values, respectively, $\bar{x}$ and $\bar{y}$ are their corresponding mean values and $s_{xy}$ is the corresponding covariance value.

%Therefore, the evaluation criterion for the  Valence-Arousal Estimation Challenge is:

\subsection{Expression Classification Challenge}\label{evaluation}

The performance measure is the average F1 Score across all 8 categories (i.e., macro F1 Score):

\begin{equation} \label{expr}
\mathcal{P}_{EXPR} = \frac{\sum_{expr} F_1^{expr}}{8}
\end{equation}

The $F_1$ score is a weighted average of the recall (i.e., the ability of the classifier to find all the positive samples) and precision (i.e., the ability of the classifier not to label as positive a sample that is negative). The $F_1$ score  takes values in the range $[0,1]$; high values are desired. The $F_1$ score is defined as:

\begin{equation} \label{f1}
F_1 = \frac{2 \times precision \times recall}{precision + recall}
\end{equation}

%Therefore, the evaluation criterion for the  Expression Classification Challenge is:

\subsection{Action Unit Detection Challenge}\label{evaluation2}

The performance measure is the average F1 Score across all 12 AUs (i.e., macro F1 Score). Therefore, the evaluation criterion for the  Action Unit Detection Challenge is:

\subsection{Emotional Reaction Intensity Estimation Challenge}\label{mtl}

The performance measure is the average Pearson’s Correlation Coefficient across the 7 emotional reactions:

\begin{equation} \label{au}
\mathcal{P}_{ERI} = \frac{\sum_{i=1}^{7} \rho^{i}}{7}
\end{equation}

\section{Baseline Networks Results} \label{baseline}

All baseline systems rely exclusively on existing open-source machine learning toolkits to ensure the reproducibility of the results. All systems have been implemented in TensorFlow. %; training time was around six hours on a Titan X GPU, with a learning rate of $10^{-4}$ and with a batch size of 256. 

In this Section, we present the baseline systems developed for each Challenge; finally we report their obtained results.

\subsection{Valence-Arousal Estimation Challenge}

\paragraph{Baseline Network} The baseline network is a ResNet with 50 layers, pre-trained on ImageNet (ResNet50) and with a (linear) output layer that gives final estimates for valence and arousal. \\

\begin{table}[ht]
\caption{Valence-Arousal Estimation Challenge Results on Validation Set: the evaluation criterion is the average CCC; CCC is displayed in \%} %on Aff-Wild and AffectNet databases}
\label{comparison_sota_va}
\centering
\begin{tabular}{ |c||c|c| }
 \hline
\multicolumn{1}{|c||}{\begin{tabular}{@{}c@{}} Teams \end{tabular}} & 
\multicolumn{1}{c|}{\begin{tabular}{@{}c@{}} CCC-V  \end{tabular}} &
\multicolumn{1}{c|}{\begin{tabular}{@{}c@{}}  CCC-A \end{tabular}} 
\\ 
  \hline
 \hline

baseline   
&  \begin{tabular}{@{}c@{}} 24.00   \end{tabular} 
& \begin{tabular}{@{}c@{}}  20.00   \end{tabular}
\\
\hline

\end{tabular}
\end{table}

\subsection{Expression Classification Challenge}

\paragraph{Baseline Network}  The baseline network is a VGG16 network with fixed (i.e., non-trainable) convolutional weights (only the 3 fully connected layers were trainable), pre-trained on the VGGFACE dataset and with an output layer equipped with softmax activation function which gives the 8 expression predictions. \\

\begin{table}[ht]
\caption{Expression Classification Challenge Results  on Validation Set: the evaluation criterion is the average F1 Score, which is displayed in \%} %on Aff-Wild and AffectNet databases}
\label{comparison_sota_expr}
\centering
\scalebox{.97}{
\begin{tabular}{ |c||c| }
 \hline
\multicolumn{1}{|c||}{\begin{tabular}{@{}c@{}} Teams \end{tabular}} & 
\multicolumn{1}{c|}{\begin{tabular}{@{}c@{}} F1  \end{tabular}} 
\\ 
  \hline
 \hline

baseline  
&  23   \\
\hline

\end{tabular}
}

\end{table}

\subsection{Action Unit Detection Challenge}

\paragraph{Baseline Network} The baseline network is a VGG16 network with  fixed convolutional weights (only the 3 fully connected layers were trained), pre-trained on the VGGFACE dataset and with an output layer equipped with sigmoid activation function which gives the 12 action unit predictions.\\

\begin{table}[ht]
\caption{Action Unit Detection Challenge Results on Validation Set: the evaluation criterion is the average F1 Score, which is displayed in \%} %on Aff-Wild and AffectNet databases}
\label{comparison_sota_au}
\centering
\scalebox{1.}{
\begin{tabular}{ |c||c| }
 \hline
\multicolumn{1}{|c||}{\begin{tabular}{@{}c@{}} Teams \end{tabular}} & 
\multicolumn{1}{c|}{\begin{tabular}{@{}c@{}} F1  \end{tabular}} 
\\ 
  \hline
 \hline

baseline 
&  39  \\
\hline

\end{tabular}
}

\end{table}

\subsection{Emotional Reaction Intensity Estimation Challenge}

The Emotional Reaction Intensity Estimation Challenge baseline results are depicted in Table~\ref{tab:reaction}. We also report the results obtained from submission to the Hume-Reaction MuSe 2022~\cite{christ2022muse} sub-challenge, as the same dataset was used. First, we observe that the audio modality provides low correlation ($.0741$), with the \ds feature set to produce better results than the eGeMAPS. This was expected as the audio is absent in several videos, making it challenging to model the modality. 

As expected, the video modality provides a higher correlation than audio, with the baseline results to obtain $.2801$ $\rho_c$ using Facial Action Units (FAU). There a number of other approaches that were submitted to MuSe 2022, but the best-performing model is obtained by the FaceRNET~\cite{kollias2023facernet}, which is comprised of a convolutional recurrent neural network with a routing mechanism on top.

Combining the audio and visual modalities does not seem to yield better results than the video models. In particular, the performance for the baseline (FAU+\ds) and the Li et al.~\cite{muse_winner} drops. Only the ViPER model seems to see performance gains of $.047$ when adding the audio modality.

\begin{table}[hbt!]
\caption{Results for emotion reaction estimation sub-challenge. The mean Pearson's Correlation Coefficient ($\rho$) for the 7 emotional reaction classes is reported, along with the confidence intervals (where possible). The baseline results for the best of 5 fixed seeds are given for each feature and late fusion configuration. The respective mean and standard deviation of the results are provided in parentheses. In addition, the approaches submitted to the MuSe 2022~\cite{christ2022muse} are presented. A dash is inserted when results are not available.}

\resizebox{1\columnwidth}{!}{
 \begin{tabular}{lcc}
 \toprule 
 & \multicolumn{2}{c}{[$\rho$]} \\
 Features & \multicolumn{1}{c}{Development} & \multicolumn{1}{c}{Test}  \\ \midrule \midrule
 \multicolumn{3}{l}{\textbf{Audio}} \\
 Baseline (eGeMAPS)~\cite{christ2022muse} & .0583 (.0504 $\pm$ .0069) & .0552 (.0479 $\pm$ .0062) \\
 Baseline (\ds)~\cite{christ2022muse} & .1087 (.0945 $\pm$ .0096) & .0741 (.0663 $\pm$ .0077) \\
 \midrule
 \multicolumn{3}{l}{\textbf{Video}} \\
 Baseline (FAU)~\cite{christ2022muse} & .2840	 (.2828 $\pm$ .0016) & .2801 (.2777 $\pm$ .0017) \\
 Baseline (\vggf)~\cite{christ2022muse} & .2488 (.2441 $\pm$ .0027) & .1830 (.1985 $\pm$ .0088) \\
 Resnet-18~\cite{muse_winner} & .3893 (----) & ---- (----) \\
 Former-DFER+MLGCN~\cite{dfer_muse_2022} & .3454 (----) & ---- (----) \\
 ViPER~\cite{viper} & .2978 (----) & .2859 (----) \\
 FaceRNET~\cite{kollias2023facernet} & \textbf{.3590} (----) & \textbf{.3607} (----) \\
 \midrule
 \multicolumn{3}{l}{\textbf{Multimodal}} \\
 Baseline~\cite{christ2022muse} & .2382	 (.2350 $\pm$ 0.0016) & .2029 (.2014 $\pm$ .0086) \\
 Resnet-18 + \ds~\cite{muse_winner} & .3968	 (----) & ---- (----) \\
 ViPER~\cite{viper} & .3025 (----) & .2970 (----) \\
 % Baseline~\cite{christ2022muse} & .2382	 (.2350 $\pm$ 0.0016) & .2029 (.2014 $\pm$ .0086) \\
 % Baseline~\cite{christ2022muse} & .2382	 (.2350 $\pm$ 0.0016) & .2029 (.2014 $\pm$ .0086) \\
 \bottomrule
 \end{tabular}\label{tab:reaction}
}
\end{table}

\section{Conclusion}\label{conclusion}

In this paper we have presented the fifth Affective Behavior Analysis in-the-wild Competition (ABAW)  held in conjunction with IEEE CVPR 2023. This Competition is a continuation of the series of ABAW Competitions.  This Competition comprises four Challenges targeting: i) uni-task Valence-Arousal Estimation, ii) uni-task Expression Classification (8 categories), iii) uni-task Action Unit Detection (12 action units) and iv) Emotional Reaction Intensity Estimation. The databases utilized for this Competition are an extended version of Aff-Wild2 and the Hume-Reaction dataset.

%%%%%%%%% REFERENCES
{\small
\bibliographystyle{ieee_fullname}
\bibliography{egbib}

\begin{thebibliography}{10}\itemsep=-1pt

\bibitem{christ2022muse}
Lukas Christ, Shahin Amiriparian, Alice Baird, Panagiotis Tzirakis, Alexander
  Kathan, Niklas M{\"u}ller, Lukas Stappen, Eva-Maria Me{\ss}ner, Andreas
  K{\"o}nig, Alan Cowen, et~al.
\newblock The muse 2022 multimodal sentiment analysis challenge: humor,
  emotional reactions, and stress.
\newblock In {\em Proceedings of the 3rd International on Multimodal Sentiment
  Analysis Workshop and Challenge}, pages 5--14, 2022.

\bibitem{cowie2000feeltrace}
Roddy Cowie, Ellen Douglas-Cowie, Susie Savvidou*, Edelle McMahon, Martin
  Sawey, and Marc Schr{\"o}der.
\newblock 'feeltrace': An instrument for recording perceived emotion in real
  time.
\newblock In {\em ISCA tutorial and research workshop (ITRW) on speech and
  emotion}, 2000.

\bibitem{kollias2022abaw}
Dimitrios Kollias.
\newblock Abaw: Valence-arousal estimation, expression recognition, action unit
  detection \& multi-task learning challenges.
\newblock {\em arXiv preprint arXiv:2202.10659}, 2022.

\bibitem{kollias2023abaw}
Dimitrios Kollias.
\newblock Abaw: Learning from synthetic data \& multi-task learning challenges.
\newblock In {\em European Conference on Computer Vision}, pages 157--172.
  Springer, 2023.

\bibitem{kollias2017recognition}
Dimitrios Kollias, Mihalis~A Nicolaou, Irene Kotsia, Guoying Zhao, and Stefanos
  Zafeiriou.
\newblock Recognition of affect in the wild using deep neural networks.
\newblock In {\em Computer Vision and Pattern Recognition Workshops (CVPRW),
  2017 IEEE Conference on}, pages 1972--1979. IEEE, 2017.

\bibitem{kollias2023facernet}
Dimitrios Kollias, Andreas Psaroudakis, Anastasios Arsenos, and Paraskeui
  Theofilou.
\newblock Facernet: a facial expression intensity estimation network.
\newblock {\em arXiv preprint arXiv:2303.00180}, 2023.

\bibitem{kollias2020analysing}
Dimitrios Kollias, Attila Schulc, Elnar Hajiyev, and Stefanos Zafeiriou.
\newblock Analysing affective behavior in the first abaw 2020 competition.
\newblock In {\em 2020 15th IEEE International Conference on Automatic Face and
  Gesture Recognition (FG 2020)(FG)}, pages 794--800. IEEE Computer Society,
  2020.

\bibitem{kollias2019face}
Dimitrios Kollias, Viktoriia Sharmanska, and Stefanos Zafeiriou.
\newblock Face behavior a la carte: Expressions, affect and action units in a
  single network.
\newblock {\em arXiv preprint arXiv:1910.11111}, 2019.

\bibitem{kollias2021distribution}
Dimitrios Kollias, Viktoriia Sharmanska, and Stefanos Zafeiriou.
\newblock Distribution matching for heterogeneous multi-task learning: a
  large-scale face study.
\newblock {\em arXiv preprint arXiv:2105.03790}, 2021.

\bibitem{kollias2018deep1}
Dimitrios Kollias, Athanasios Tagaris, Andreas Stafylopatis, Stefanos Kollias,
  and Georgios Tagaris.
\newblock Deep neural architectures for prediction in healthcare.
\newblock {\em Complex \& Intelligent Systems}, 4(2):119--131, 2018.

\bibitem{kollias2019expression}
Dimitrios Kollias and Stefanos Zafeiriou.
\newblock Expression, affect, action unit recognition: Aff-wild2, multi-task
  learning and arcface.
\newblock {\em arXiv preprint arXiv:1910.04855}, 2019.

\bibitem{kollias2021affect}
Dimitrios Kollias and Stefanos Zafeiriou.
\newblock Affect analysis in-the-wild: Valence-arousal, expressions, action
  units and a unified framework.
\newblock {\em arXiv preprint arXiv:2103.15792}, 2021.

\bibitem{kollias2021analysing}
Dimitrios Kollias and Stefanos Zafeiriou.
\newblock Analysing affective behavior in the second abaw2 competition.
\newblock In {\em Proceedings of the IEEE/CVF International Conference on
  Computer Vision}, pages 3652--3660, 2021.

\bibitem{muse_winner}
Jia Li, Ziyang Zhang, Junjie Lang, Yueqi Jiang, Liuwei An, Peng Zou, Yangyang
  Xu, Sheng Gao, Jie Lin, Chunxiao Fan, Xiao Sun, and Meng Wang.
\newblock Hybrid multimodal feature extraction, mining and fusion for sentiment
  analysis.
\newblock In {\em Proceedings of the 3rd International on Multimodal Sentiment
  Analysis Workshop and Challenge}, MuSe' 22, page 81–88, New York, NY, USA,
  2022. Association for Computing Machinery.

\bibitem{viper}
Lorenzo Vaiani, Moreno La~Quatra, Luca Cagliero, and Paolo Garza.
\newblock Viper: Video-based perceiver for emotion recognition.
\newblock In {\em Proceedings of the 3rd International on Multimodal Sentiment
  Analysis Workshop and Challenge}, page 67–73, New York, NY, USA, 2022.
  Association for Computing Machinery.

\bibitem{dfer_muse_2022}
Kexin Wang, Zheng Lian, Licai Sun, Bin Liu, Jianhua Tao, and Yin Fan.
\newblock Emotional reaction analysis based on multi-label graph convolutional
  networks and dynamic facial expression recognition transformer.
\newblock In {\em Proceedings of the 3rd International on Multimodal Sentiment
  Analysis Workshop and Challenge}, page 75–80, New York, NY, USA, 2022.
  Association for Computing Machinery.

\bibitem{zafeiriou2017aff}
Stefanos Zafeiriou, Dimitrios Kollias, Mihalis~A Nicolaou, Athanasios
  Papaioannou, Guoying Zhao, and Irene Kotsia.
\newblock Aff-wild: Valence and arousal ‘in-the-wild’challenge.
\newblock In {\em Computer Vision and Pattern Recognition Workshops (CVPRW),
  2017 IEEE Conference on}, pages 1980--1987. IEEE, 2017.

\end{thebibliography}
}

\end{document}